\documentclass{CVM}

\CVMsetup{
type      = {Research/Review Article},
doi       = {CVM.XXXX},
title     = {Disentangled Geometry and Appearance for Efficient Multi-View Surface Reconstruction and Rendering},
author    = {Qitong Zhang$^{1}$ and Jieqing Feng$^{1}$\cor{}\\
},
runauthor = {Q. Zhang, J. Feng},
abstract  = {
  This paper addresses the limitations of neural rendering-based multi-view surface reconstruction methods, which require an additional mesh extraction step that is inconvenient and would produce poor-quality surfaces with mesh aliasing, restricting downstream applications. Building on the explicit mesh representation and differentiable rasterization framework, this work proposes an efficient solution that preserves the high efficiency of this framework while significantly improving reconstruction quality and versatility. Specifically, we introduce a disentangled geometry and appearance model that does not rely on deep networks, enhancing learning and broadening applicability. A neural deformation field is constructed to incorporate global geometric context, enhancing geometry learning, while a novel regularization constrains geometric features passed to a neural shader to ensure its accuracy and boost shading. For appearance, a view-invariant diffuse term is separated and baked into mesh vertices, further improving rendering efficiency. Experimental results demonstrate that the proposed method achieves state-of-the-art training (4.84 minutes) and rendering (0.023 seconds) speeds, with reconstruction quality that is competitive with top-performing methods. Moreover, the method enables practical applications such as mesh and texture editing, showcasing its versatility and application potential. This combination of efficiency, competitive quality, and broad applicability makes our approach a valuable contribution to multi-view surface reconstruction and rendering.
},
keywords  = {multi-view 3D reconstruction; surface rendering; neural deformation field; neural shader},
copyright = {The Author(s)},
}

\begin{document}

\maketitle

\enlargethispage{-3pt}
\begin{figure}[b] \vskip -4mm
\small\renewcommand\arraystretch{1.3}
\begin{tabular}{p{80.5mm}} \toprule\\ \end{tabular}
\vskip -4.5mm \noindent \setlength{\tabcolsep}{1pt}
\begin{tabular}{p{3.5mm}p{80mm}}
$1\quad $ & State Key Lab of CAD \& CG, Zhejiang University, Hangzhou 310058, China. E-mail: Q. Zhang, zhangqitong@zju.edu.cn; J. Feng, jqfeng@cad.zju.edu.cn\cor{}.\\
&\hspace{-5mm} Manuscript received: 2025-02-04; accepted: 202x-xx-xx\vspace{-2mm}
\end{tabular} \vspace {-3mm}
\end{figure}

\section{Introduction}

Recovering 3D shapes from multi-view images is essential for applications in computer vision and graphics, including games, 3D movies, animation, and VR/AR, etc. In contrast to traditional reconstruction methods~\cite{1, 2, 3, 4, 5} that rely on accurate feature matching between images, the methods based on inverse rendering~\cite{6, 7, 8, 9, 10, 11} present a promising alternative due to their capacity to handle complex cases such as non-Lambertian surfaces or thin structures. Geometry and appearance can be optimized in a self-supervised manner by matching differentiable rendered images with captured images.

Among inverse rendering-based methods, neural rendering-based approaches represent geometry implicitly using neural networks, such as \textit{multi-layer perceptrons} (MLPs). These methods have achieved impressive results with representations like density fields~\cite{7, 12, 13, 14, 15, 16, 17} or \textit{signed distance fields} (SDF)~\cite{8, 9, 11}, combined with volume rendering techniques for high-quality surface reconstruction. However, the reliance on large neural networks and ray marching significantly limits training and rendering efficiency. While some variants~\cite{16, 18} improve efficiency, these methods still require additional mesh extraction operations (e.g., Marching Cubes~\cite{19}) that are inconvenient and may lead to low-quality meshes with aliasing, particularly at low triangle counts, that negatively impact downstream applications.

In contrast, methods leveraged explicit representations to model geometry reduce reliance on large neural networks and enable efficient training and rendering. Among them, mesh-based methods are widely adopted for their compatibility with standard graphics workflows, facilitating integration into downstream applications.  Recently, some differentiable rasterization-based methods~\cite{20, 21, 22, 23} have gained attention for their ability to simulate the interaction of arbitrary lighting, materials, and geometry without requiring priors, broadening their applicability. However, challenges remain in balancing mesh quality and training efficiency. Some methods suffer from limited geometry learning capabilities~\cite{20}, while others require additional pre-trained models~\cite{21} or combine implicit representations~\cite{22, 23} during training, leading to increased computational costs.

To overcome these limitations, this paper proposes an efficient and versatile differentiable rasterization-based method that significantly improves reconstruction quality, maintains high efficiency, and expands application potential. The key innovation lies in a disentangled geometry and appearance framework, which reduces optimization complexity while enhancing their respective learning capabilities. For geometry, a neural deformation field improves geometry learning by modeling continuous vertex deformations and geometric features, which are transferred to a neural shader via rasterization to enhance shading. For appearance, a baking strategy inspired by~\cite{24, 25, 26, 27} separates and stores view-invariant diffuse colors in mesh vertices to accelerate rendering by simplifying neural shader computations to view-dependent specular predictions. Furthermore, a novel regularization is incorporated into the framework to constrain learned geometric features and ensure robust and accurate shading. 

The proposed method supports diverse applications, such as refinement, attribute transfer, mesh morphing and adaptation, making it highly versatile for real-time scenarios. By decoupling geometry and appearance, attributes like shape and texture can be seamlessly transferred between models. The neural deformation field facilitates smooth interpolation and morphing of topologically consistent meshes, facilitating animations and modifications. Additionally, the fast rendering capability significantly enhances its practicality in real-time scenarios, such as interactive editing and visual effects.

Experiments show that the proposed method achieves state-of-the-art training (4.84 min) and rendering (0.023 s) speeds, and competitive reconstruction quality with top-performing methods, all without requiring pre-trained models or post-processing. Extensive experiments further validate the effectiveness of this method across various editing tasks. The main contributions are summarized as follows:

\begin{itemize}

  \item A disentangled geometry and appearance framework that enhances learning capabilities and enables versatile editing tasks such as attributes transfer.
  
  \item A neural deformation field that learns continuous vertex deformations and geometric features, improving geometry learning and supporting applications like mesh modification and morphing.
  
  \item An efficient baking strategy that accelerates rendering by storing view-invariant diffuse colors in mesh vertices while ensuring shading robustness through geometric feature regularization, supporting real-time applications.

\end{itemize}

\section{Related Work}

\textbf{Classical Methods} for multi-view reconstruction are typically based on \textit{multi-view stereo} (MVS), which relies on minimizing the matching cost of features across images to find accurate correspondence. Although traditional MVS methods~\cite{1, 2, 3} using hand-crafted features can achieve high accuracy, they struggle in low-textured regions, limiting reconstruction completeness. Learning-based MVS approaches~\cite{4, 5, 28} improve the correspondence matching by introducing semantic information embedded in learned features but require 3D supervision. Additionally, the above methods involve multi-stage pipelines with post-processing steps, such as depth fusion and Poisson surface reconstruction~\cite{29}, to obtain the final mesh. Unlike these methods, the proposed method directly deforms an initial mesh to the target mesh without requiring strict correspondence constraints or additional steps, leveraging differentiable rendering techniques for self-supervised learning and better handling of non-Lambertian surfaces.

\textbf{Neural Rendering-Based Methods} optimize virtual scene parameters encoded by implicit functions, typically through training a large neural network. Early approaches~\cite{30, 31, 32, 33} modeled implicit functions using MLPs to represent SDFs~\cite{30}, occupancy fields~\cite{31}, or binary indicators~\cite{32, 33}, requiring 3D datasets for training. Recent advancements in neural rendering avoid the 3D supervision by using differentiable rendering techniques to match the appearance of rendered views with input camera images, enabling end-to-end self-supervised training. These methods generally fall into two categories based on rendering techniques: surface rendering and volume rendering. Surface rendering approaches, such as IDR~\cite{10} and DVR~\cite{34}, seek the intersection between the rays and the underlying surface represented by an SDF or occupancy field, assuming ray color is determined solely by the intersection point. Volume rendering approaches, like NeRF~\cite{7} and its variants~\cite{12, 13, 14, 15, 16, 17} sample multiple points along each ray and predict color and density at each point to synthesize the radiance field, achieving notable results in novel view synthesis. However, surfaces derived from density fields often suffer from noise due to undefined boundaries. NeuS~\cite{9} and VolSDF~\cite{11} address this problem by replacing density field with SDF for volume rendering, yielding more accurate surface reconstructions. Some subsequent works~\cite{35, 36, 37} further enhance the geometry details. Despite these advances, the reliance on ray marching for rendering incurs high computational costs during both training and inference, limiting efficiency. To enhance efficiency, multi-resolution hash encoding~\cite{16} has been introduced to accelerate training in SDF networks~\cite{18, 38}. However, integrating these methods into traditional graphics pipelines still requires post-processes like Marching Cubes~\cite{19}, which may produce low-quality meshes unsuitable for downstream applications. The proposed method avoids these limitations by combining explicit mesh representations with a neural shader through a differentiable renderer, eliminating surface extraction while achieving fast training and high-quality reconstructions.

\textbf{Explicit Representations} have been explored to improve reconstruction efficiency and quality. Point clouds are used in~\cite{39} to construct a point-based radiance field that jointly optimizes MLPs and point clouds, but require complex processing steps, such as MVS depth fusion. Voxel-based approaches~\cite{40, 41, 42, 43} accelerate training by substituting MLPs with voxel grids, though their resolution limits the recovery of fine details. Recent works like 3D Gaussian Splatting~\cite{44}, which represents 3D objects using Gaussian distributions, achieve excellent real-time performance but struggle to extract high-quality native geometry.

With advances in differentiable rendering~\cite{45}, many recent methods directly optimize mesh geometry and appearance to reconstruct 3D objects, compatible with traditional graphics pipeline. Differentiable path tracing methods~\cite{46, 47, 48} jointly optimize geometry, materials and lighting parameters from images but rely on prior assumptions about lighting and material models~\cite{49, 50}, which limits their generalization to arbitrary scenes. Other methods bypass explicit light transport modeling using differentiable rasterization and deferred shading to optimize meshes. For example, NVDIFFREC~\cite{51} and NVDIFFRECmc~\cite{52} render the mesh extracted from implicit SDFs with DMTet~\cite{53} using the differentiable rasterizer Nvdiffrast~\cite{54}. Despite their efficiency, these methods are limited by the uneven triangle distribution in tetrahedral meshes and simplified shading models. Neural shader-based approaches enhance reconstruction flexibility by fully parameterizing shading models for arbitrary lighting and materials. NDS~\cite{20} directly optimizes vertex offsets through a neural shader but lacks effective geometric learning mechanisms. FastMESH~\cite{21} and CPT-VR~\cite{23} incorporate pre-trained MVS models~\cite{28} as geometric priors, improving reconstruction accuracy at the cost of additional constraints and pretraining overhead. Some methods~\cite{22, 23} integrate implicit SDF fields to generate initial meshes for improved surface accuracy, though the repeated mesh extraction during training significantly slows down the process. In contrast, our method introduces a neural deformation field to enrich mesh geometric details without requiring external priors or iterative surface extraction, ensuring efficient training. Additionally, by adopting a baking strategy like~\cite{24, 25, 26} to store view-invariant diffuse appearance directly in mesh vertices, our method minimizes MLPs computations in appearance learning, further optimizing efficiency for real-time applications.

\begin{figure*}[t]
  \centering
  \includegraphics[width=\linewidth]{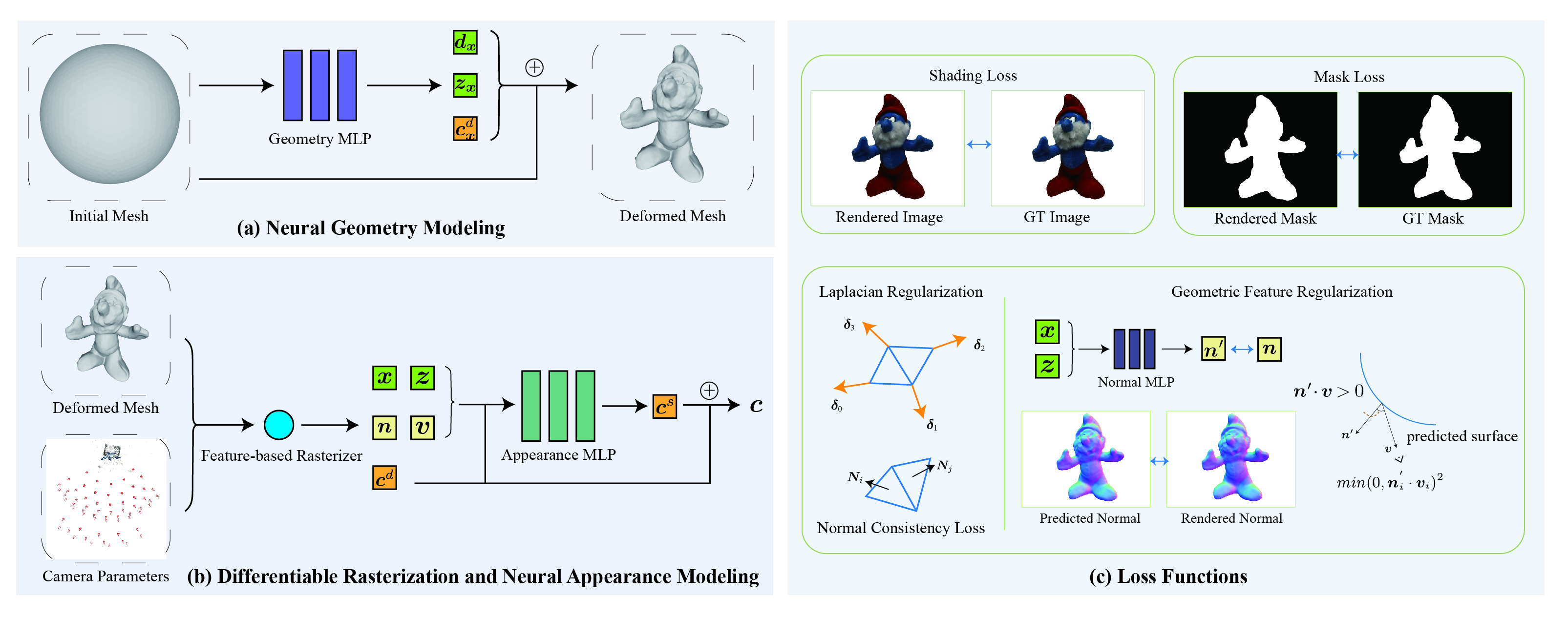}
  \caption{Method Overview. In (a), the deformation vectors $\boldsymbol{d}_{\boldsymbol{x}}$ and geometric features $\boldsymbol{z}_{\boldsymbol{x}}$ of mesh vertices $\boldsymbol{x}$ are added to the initial mesh to obtain a deformed mesh. The view-independent diffuse components $\boldsymbol{c}^d_{\boldsymbol{x}}$ are stored at the mesh vertices. In (b), the rasterizer outputs, including the position map $\boldsymbol{x}$, geometric feature map $\boldsymbol{z}$, normal map $\boldsymbol{n}$, and diffuse color map $\boldsymbol{c}^d$, are stored in the g-buffer, combining with the view direction $\boldsymbol{v}$ to input into an appearance MLP to predict the specular component $\boldsymbol{c}^s$. The final image $\boldsymbol{c}$ is obtained as $\boldsymbol{c} = \boldsymbol{c}^{d} + \boldsymbol{c}^{s}$. In (c), the loss function comprises shading loss, mask loss, Laplacian regularization, normal consistency loss, and geometric feature regularization. Here, $\boldsymbol{\delta}_i$, $\boldsymbol{N}_i$ and $\boldsymbol{N}_j$ denote the Laplacian coordinates and the normals of triangular mesh faces, respectively. Geometric feature regularization ensures that the predicted normals $\boldsymbol{n}^{\prime}$, obtained via a normal MLP network that inputs positions and geometric features, align with the rasterized normals $\boldsymbol{n}$, thereby maintaining accurate geometric features. Additionally, the constraint $\boldsymbol{n}^{\prime}_{i} \cdot \boldsymbol{v}_i > 0$ ensures the predicted normals face the camera.}
  \label{fig2}
\end{figure*}

\section{Method}
Given a set of images $\mathcal{I}=\left\{I_{1}, I_{2}, \ldots, I_{n}\right\}$ with known camera parameters and corresponding masks $\mathcal{M}=\left\{M_{1}, M_{2}, \ldots, M_{n}\right\}$, our goal is to obtain high-quality 3D surface and photo-realistic novel views of an object with efficient training and inference. The whole pipeline of the proposed method is depicted in Fig.~\ref{fig2}. First, the preliminaries of mesh-based rendering are introduced in Section 3.1. Then, the neural geometry model with a deformation field, feature-based differentiable rasterization, and the neural appearance model based on a baking strategy are presented in Sections 3.2, 3.3, and 3.4, respectively. Finally, Section 3.5 describes the components of loss functions.

\subsection{Preliminaries}
To efficiently recover 3D content compatible with traditional graphics pipeline, the proposed method follows the mesh-based rendering framework to optimize geometry and appearance.

\textbf{Initial Mesh.} The surface is represented by a triangle mesh $\mathcal{G}=(\mathcal{V}, \mathcal{E}, \mathcal{F})$, where $\mathcal{V}$ denotes vertices, $\mathcal{E}$ edges, and $\mathcal{F}$ faces. The proposed method deforms the triangle mesh gradually using gradient descent, maintaining topological invariance. Consequently, the quality of the initial mesh significantly impacts the final mesh quality. Inspired by~\cite{20, 51}, we considered two types of initial meshes: a visual hull computed from masks and a tetrahedral mesh extracted from implicit SDFs using DMTet~\cite{53}. The visual hull suffers from self-intersections and holes, leading to low mesh quality, while the tetrahedral mesh is limited by triangle inhomogeneity and a time-consuming extraction process. To avoid these issues, we use an icosahedral mesh on the unit sphere as the initial mesh, which has a uniform triangle distribution. During training, the deformed mesh is upsampled to achieve higher resolution and improve mesh quality. 

\textbf{Rendering of Explicit Surface.} \textit{Physically-based rendering} (PBR) is commonly used to generate high-quality, realistic images by modeling physical properties. Let $\boldsymbol{x}$ denote the intersection point between a ray $\boldsymbol{r}$ along the view direction $\boldsymbol{v}$ and the explicit surface. The radiance at $x$ in the direction $\boldsymbol{w}_{o}=-\boldsymbol{v}$ is computed by the non-emissive rendering equation:
\begin{equation}
  L(\boldsymbol{x},\boldsymbol{w}_{o}) = \int_{\Omega}L_{i}(\boldsymbol{x},\boldsymbol{w}_{i}) f(\boldsymbol{x},\boldsymbol{w}_{i},\boldsymbol{w}_{o}) (\boldsymbol{w}_{i}\cdot\boldsymbol{n}) d\boldsymbol{w}_{i},
\label{eq1}
\end{equation}
where $L_{i}$ denotes the incident radiance of $\boldsymbol{x}$ from the incoming direction $\boldsymbol{w}_{i}$, $f$ is the bidirectional reflectance distribution function (BRDF), which describes the proportion of reflected radiance in the direction $\boldsymbol{w}_{o}$ with respect to incoming radiance from the direction $\boldsymbol{w}_{i}$. The integration is performed over all incident directions $\boldsymbol{w}_{i}$ on the hemisphere $\Omega$ where $\boldsymbol{w}_{i} \cdot \boldsymbol{n} > 0$, and is typically approximated using finite sampling to reduce computational cost. Works such as~\cite{10, 20} leverage neural networks to approximate PBR, constructing an MLP-based neural shader that can handle arbitrary illuminations and materials, efficiently rendering images as follows:
\begin{equation}
  C(\boldsymbol{r}) = h_{\boldsymbol{\sigma}}(\boldsymbol{x},\boldsymbol{n},\boldsymbol{v}),
\label{eq2}
\end{equation}
where the neural shader $h_{\boldsymbol{\sigma}}$ with parameters $\boldsymbol{\sigma}$ takes as input the surface intersection point $\boldsymbol{x}$, the surface normal $\boldsymbol{n}$, and the view direction $\boldsymbol{v}$, and outputs the shading RGB colors $C(\boldsymbol{r})$ of the ray $\boldsymbol{r}$.

\subsection{Neural Geometry Modeling}
Although vertex offsets of the initial mesh can be directly optimized using the neural shader, relying solely on the local geometry of the vertices limits the accuracy of the final mesh. To overcome this problem, a neural deformation field is introduced to capture and incorporate global geometric context, thereby improving the learning capacity and facilitating the recovery of finer surface details. A mapping $g\colon\mathbb{R}^{3}\to\mathbb{R}^{3}$ is constructed through an MLP, which maps a continuous domain $S$ to another continuous domain $T$. The mapping is applied to the initial mesh vertices $\mathcal{V}\in S$ to obtain the deformed mesh vertices $\mathcal{V}^{\prime}\in T$. The deformation MLP $g_{\boldsymbol{\theta}}$ parameterized by weights $\boldsymbol{\theta}$ is formulated as follows:
\begin{equation}
  (\boldsymbol{d}_{\boldsymbol{x}},\boldsymbol{z}_{\boldsymbol{x}}) = g_{\boldsymbol{\theta}}(\boldsymbol{x}, e_{\boldsymbol{\varphi}}(\boldsymbol{x})),
\label{eq3}
\end{equation}
where the network inputs are the vertices $\boldsymbol{x}$ enhanced by a multi-resolution hash encoding $e_{\boldsymbol{\varphi}}(\boldsymbol{x})$, and the network outputs are the deformation vectors $\boldsymbol{d}_{\boldsymbol{x}}$ and the geometric features $\boldsymbol{z}_{\boldsymbol{x}}$. We apply the deformation MLP to the initial mesh vertices $\mathcal{V}$, compute the deformation vectors $\boldsymbol{d}_{\mathcal{V}}$ and features $\boldsymbol{z}_{\mathcal{V}}$, and finally, obtain the deformed mesh vertices $\mathcal{V}^{\prime} = \mathcal{V} + \boldsymbol{d}_{\mathcal{V}}$. The deformed triangle mesh is denoted as $\mathcal{G}^{\prime}=(\mathcal{V}^{\prime}, \mathcal{E}, \mathcal{F}, \mathcal{Z})$, where $\mathcal{Z}$ represents the geometric features associated with the vertices $\mathcal{V}^{\prime}$ and can be passed to subsequent stages to enhance learning.

\subsection{Differentiable Rasterization}
After obtaining the deformed mesh, a differentiable mesh renderer   based on the deferred shading pipeline, widely used in real-time graphics applications, is employed to generate 2D images from each view for subsequent appearance estimation. The first step is to render the triangle mesh from known camera views using a feature-based differentiable rasterizer, Nvdiffrast~\cite{54}. Specifically, for each pixel $p(x,y)$ in the projection of the triangle face $(\boldsymbol{x}_{A}, \boldsymbol{x}_{B}, \boldsymbol{x}_{C})$ of the mesh, we compute the barycentric coordinates $(u, v)$, depth $D_{p}$, visibility mask $M_{p}$ using the rasterizer as follows:
\begin{equation}
  (u, \nu, D_{p}, M_{p}) = \mathcal{R}(\mathcal{G}, p(x, y), K, R, \boldsymbol{t}),
\label{eq4}
\end{equation}
where $\mathcal{R}$ is the rasterizer, and $K, R, \boldsymbol{t}$ are the camera’s intrinsic and extrinsic parameters, respectively. Next, the position $\boldsymbol{x}_{p}$, normal $\boldsymbol{n}_{p}$, and geometric feature $\boldsymbol{z}_{p}$ of each pixel $p(x, y)$ projected onto the mesh are interpolated linearly using the barycentric coordinates $(u, v)$ and the corresponding attributes of each triangle vertex:
\begin{equation}
\begin{footnotesize}
  (\boldsymbol{x}_{p}\quad\boldsymbol{n}_{p}\quad\boldsymbol{z}_{p}) = (u\quad v\quad1-u-v)\begin{pmatrix}\boldsymbol{x}_{A}&\boldsymbol{n}_{A}&\boldsymbol{z}_{A}\\\boldsymbol{x}_{B}&\boldsymbol{n}_{B}&\boldsymbol{z}_{B}\\\boldsymbol{x}_{C}&\boldsymbol{n}_{C}&\boldsymbol{z}_{C}\end{pmatrix}.
\end{footnotesize}
\label{eq5}
\end{equation}
The position map, normal map, and geometric feature map are constructed and stored in the geometry buffer (g-buffer) for use in subsequent stages.

\subsection{Neural Appearance Modeling}
At this stage, we employ an MLP network for appearance modeling, simulating the interaction between geometry, material, and light using information stored in the g-buffer to output RGB colors. Differently, a baking strategy inspired by~\cite{24, 25, 27} is adopted to further reduce MLP forward passes and improve rendering efficiency. The output RGB color is decoupled into diffuse and specular components. We first replace the MLP prediction by pre-computing and storing, i.e. \textit{baking}, the diffuse color in the deformed mesh vertices. Specifically, the diffuse color is view-invariant and thus can be learned by mapping from 3D points $\boldsymbol{x}$ to their diffuse colors  $\boldsymbol{c}_{\boldsymbol{x}}^{d}$. We integrate this process into the geometry MLP $g_{\boldsymbol{\theta}}$, compute the diffuse colors $(\boldsymbol{c}_{\boldsymbol{A}}^{d},\boldsymbol{c}_{\boldsymbol{B}}^{d},\boldsymbol{c}_{\boldsymbol{C}}^{d})$ of the triangle vertices, and render the diffuse color maps for known camera views via differentiable rasterization and store them in the g-buffer. Consequently, the Equation ~\eqref{eq3} and ~\eqref{eq5} are updated as follows:
\begin{equation}
  (\boldsymbol{d}_{\boldsymbol{x}},\boldsymbol{z}_{\boldsymbol{x}},\boldsymbol{c}_{\boldsymbol{x}}^{d}) = g_{\boldsymbol{\theta}}(\boldsymbol{x},e_{\boldsymbol{\varphi}}(\boldsymbol{x})),
\label{eq6}
\end{equation}
\begin{equation}
\begin{scriptsize}
  (\boldsymbol{x}_p\hspace{0.5em}\boldsymbol{n}_p\hspace{0.5em}\boldsymbol{z}_p\hspace{0.5em}\boldsymbol{c}_p^d) = (u\hspace{0.5em}v\hspace{0.5em}1-u-v)\renewcommand{\arraystretch}{1.5}\setlength{\arraycolsep}{3pt} \begin{pmatrix}\boldsymbol{x}_A&\boldsymbol{n}_A&\boldsymbol{z}_A&\boldsymbol{c}_A^d\\\boldsymbol{x}_B&\boldsymbol{n}_B&\boldsymbol{z}_B&\boldsymbol{c}_B^d\\\boldsymbol{x}_C&\boldsymbol{n}_C&\boldsymbol{z}_C&\boldsymbol{c}_C^d\end{pmatrix}.
\end{scriptsize}
\label{eq7}
\end{equation}
Then, the g-buffer is processed by an MLP-based neural shader, which leverages the geometric feature $\boldsymbol{z}$ to enhance learning capacity and predicts only the specular color $\boldsymbol{c}^{s}$. The Equation ~\eqref{eq2} is modified as follows:
\begin{equation}
  \boldsymbol{c}^{s} = h_{\boldsymbol{\sigma}}(\boldsymbol{x},\boldsymbol{z},\boldsymbol{n}, SH(\boldsymbol{v}), \boldsymbol{c}^{d}),
\label{eq8}
\end{equation}
where $\boldsymbol{c}^{d}$ represented the base (diffuse) color, and the MLP predicts the view-dependent residual based on the Spherical Harmonics encoded view direction $SH(\boldsymbol{v})$. Finally, the color $\boldsymbol{c}$ is obtained as $\boldsymbol{c} = \boldsymbol{c}^{d} + \boldsymbol{c}^{s}$.

\subsection{Loss Functions}
During training, the optimization in the proposed method is implemented by minimizing the following loss function:
\begin{equation}
  \min_{\boldsymbol{\theta}, \boldsymbol{\sigma}, \boldsymbol{\eta}}L_{a}(\mathcal{G}, \boldsymbol{\theta}, \boldsymbol{\sigma};\mathcal{I}, \mathcal{M}) + L_{g}(\mathcal{G}, \boldsymbol{\theta}, \boldsymbol{\eta};\mathcal{I}, \mathcal{M}),
\label{eq9}
\end{equation}
where $L_{a}$, $L_{g}$ are the appearance loss and geometry regularization loss, respectively.

\textbf{Appearance Loss.} It comprises a shading loss $L_{shading}$ and a mask loss $L_{mask}$, weighted by $\lambda_{s}$ and $\lambda_{m}$, respectively:
\begin{equation}
  L_{a} = \lambda_{s}L_{shading} + \lambda_{m}L_{mask},
\label{eq10}
\end{equation}
\begin{equation}
  L_{shading} = \frac{1}{|\mathcal{I}|}\sum_{i = 1}^{|\mathcal{I}|}\|I_{i} - \overline{I_{i}}\|_{1},
\label{eq11}
\end{equation}
\begin{equation}
  L_{mask} = \frac{1}{|\mathcal{M}|}\sum_{i = 1}^{|\mathcal{M}|}\|M_{i} - \overline{M_{i}}\|_{1},
\label{eq12}
\end{equation}
where $L_{shading}$ ensures the rendered color images $\overline{I_{i}}$ match the real color images $I_{i}$, and $L_{mask}$ ensures the rendered masks $\overline{M_{i}}$ match the target masks $M_{i}$. During each iteration, a random subset of pixels from the intersection between the input and rendered masks is sampled, and the L1 losses $\|.\|_{1}$ are computed.

\textbf{Geometry Regularization.} To prevent undesirable degenerated triangles and self-intersections, and ensure mesh smoothness, we adopt Laplacian regularization loss $L_{laplacian}$ and normal consistency loss $L_{normal}$, similar to~\cite{20}:
\begin{equation}
  L_{laplacian} = \frac{1}{n}\sum_{i = 1}^{n}\|\boldsymbol{\delta}_{i}\|_{2}^{2} ,\quad\boldsymbol{\delta}_{i} = (LV)_{i},
\label{eq13}
\end{equation}
\begin{equation}
  L_{normal} = \frac{1}{|\overline{\mathcal{F}}|}\sum_{(i, j) \in \overline{\mathcal{F}}}(1-\boldsymbol{N}_{i}\cdot\boldsymbol{N}_{j})^{2}.
\label{eq14}
\end{equation}
where $L$ is the graph Laplacian of the mesh, $V$ is a matrix with vertex position as rows, $\boldsymbol{\delta}_{i}$ is the differential coordinates~\cite{55} of vertex $i$, $\|.\|_{2}$ is the Euclidean norm, and $\boldsymbol{N}_{i}$ and $\boldsymbol{N}_{j}$ are normals of adjacent triangle faces in the neighboring triangle pair set $\overline{\mathcal{F}}$.

Additionally, we introduce a novel geometric feature regularization strategy to enhance the expressiveness of geometric features. A small MLP, denoted as $p_{\boldsymbol{\eta}}$ and parameterized by $\boldsymbol{\eta}$, is designed to predict point normals $\boldsymbol{n}^{\prime}$ from the input positions $\boldsymbol{x}$ and their geometric features $\boldsymbol{z}$:
\begin{equation}
  \boldsymbol{n}^{\prime} = p_{\boldsymbol{\eta}}(\boldsymbol{x},\boldsymbol{z}).
\label{eq15}
\end{equation}
The geometric features are optimized by matching the predicted normals $\boldsymbol{n}^{\prime}_{i}$ with the rendered normals $\boldsymbol{n}_{i}$ obtained from the rasterizer. To ensure consistency, the predicted normal is constrained to face the camera, such that $\boldsymbol{n}^{\prime}_{i}\cdot \boldsymbol{v}_i > 0$. The geometric feature regularization loss is defined as:
\begin{small}
\begin{equation}
  L_{feature} = \frac{1}{|\mathcal{I}|}\sum_{i = 1}^{|\mathcal{I}|} (MSE(\boldsymbol{n}_{i}, \boldsymbol{n}_{i}^{'}) + min(0, \boldsymbol{n}_{i}^{'}\cdot\boldsymbol{v}_{i})^{2}),
\label{eq16}
\end{equation}
\end{small}
where the mean squared error (MSE) of these two normals is computed. The complete geometric regularization loss is then formulated as:
\begin{equation}
  L_{g} = \lambda_{l}L_{laplacian} + \lambda_{n}L_{normal} + \lambda_{f}L_{feature},
\label{eq17}
\end{equation}
where $\lambda_{l}$, $\lambda_{n}$, and $\lambda_{f}$ are the respective weights for the loss terms.

\section{Experiments and Disussions}
In this section, we first describe the implementation details of the proposed method. Then, the effectiveness of the proposed method is validated, and the quantitative and qualitative evaluations are done by comparing it with other state-of-the-art methods. After that, we demonstrate applications of the proposed methods, such as texture transfer and mesh editing including deformation, morphing, and refinement. Finally, we discuss the parameter analysis, superiority, and limitation of the proposed method.

\subsection{Implementation Details and Setup}
The proposed method is implemented in the PyTorch~\cite{56} framework with CUDA extensions. All experiments are carried out on a PC with an Intel Core I7-6700K CPU, 64GB RAM, and a single GeForce RTX 3090 GPU.

\begin{table*}[t]
  \centering
  \caption{Quantitative results of ablation studies on DTU dataset. TT and RT deonotes training time and rendering time, respectively.}
  \setlength{\tabcolsep}{1mm}{
    \begin{tabular}{c|c|c|c|c|c|c|c|c|c}
    \hline
    \multirow{2}{*}{Settings} & \multicolumn{5}{c|}{Design Choices} & \multirow{2}{*}{CD$\downarrow$} & \multirow{2}{*}{PSNR$\uparrow$} & \multirow{2}{*}{TT(m)$\downarrow$} & \multirow{2}{*}{RT(s)$\downarrow$} \\
    \cline{2-6}
    
      & Geometry MLP & Geometric Feature & Diffuse Baking & Appearance MLP & $L_{feature}$ &  &  &  & \\
    \hline
    
      (a) &  &  &  & \checkmark &  & 6.46 & 23.08 & \textbf{2.85} & 0.043 \\
        
      (b) & \checkmark &  & \checkmark & \checkmark &  & 1.80 & 29.07 & 2.99 & 0.029 \\

      (c) & \checkmark & \checkmark & \checkmark & \checkmark &  & 1.49 & 29.45 & 3.33 & 0.023 \\

      (d) & \checkmark & \checkmark & \checkmark & \checkmark & \checkmark & 1.34 & \textbf{30.04} & 4.84 & \textbf{0.023} \\

      (e) & \checkmark & \checkmark &  & \checkmark & \checkmark & \textbf{1.32} & 29.80 & 5.22 & 0.038 \\
    \hline
    \end{tabular}
  }
  \label{tab1}
\end{table*}

\begin{figure}[t]
  \centering
  \includegraphics[width=\linewidth]{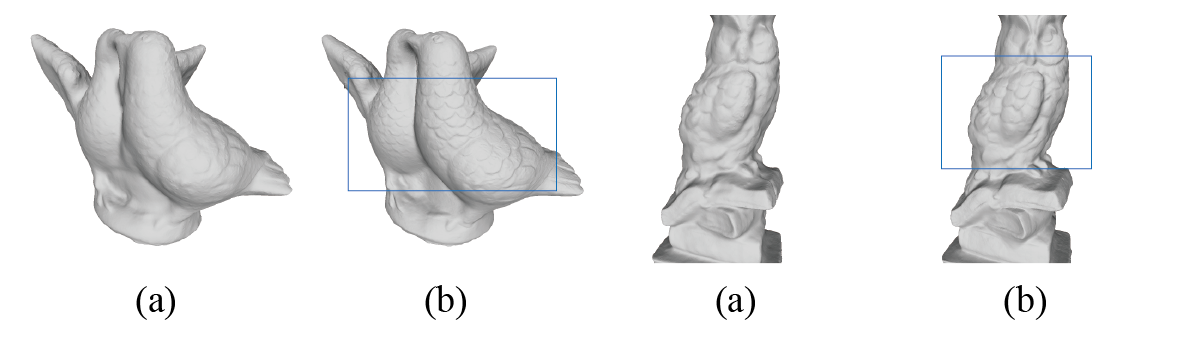}
  \caption{Meshes obtained by using different upsampling strategy. (a) NDS upsamling. (b) Our upsamling.}
  \label{fig3}
\end{figure}

\textbf{Optimization.} We adopt the Adam~\cite{57} optimizer for momentum-based gradient descent. Similar to previous works~\cite{20, 58}, we use a coarse-to-fine scheme for robust training. We train the proposed method for 2000 iterations, randomly selecting one view to compute appearance, and shading $75\%$ of mask pixels at every iteration. A coarse initial icosahedral mesh with 2562 vertices is used. After 500 iterations, the current mesh is upsampled and finally subdivided into 163842 vertices using loop subdivision~\cite{59}. The upsampling strategy used in NDS~\cite{20}, which upsamles the mesh every 500 iterations, was also tested. As shown in Fig.~\ref{fig3}, compared with our upsampling strategy, the surface details of meshes reconstructed by this strategy is not as salient as the proposed method. We use a learning rate of $2\cdot10^{-3}$ for geometry neural network decreasing by $25\%$ every 500 iterations, and a learning rate of $1\cdot10^{-3}$ for appearance modeling. Similar to~\cite{20}, the loss weights of individual terms $\lambda_{s}$, $\lambda_{m}$, $\lambda_{l}$, $\lambda_{n}$, $\lambda_{f}$ are set to 1.0, 2.0, 40, 0,01, 0,1, respectively. We also increase the weights of the regularization terms $\lambda_{l}$, $\lambda_{n}$, $\lambda_{f}$ by a factor of 64.

\begin{figure}[b]
  \centering
  \includegraphics[width=\linewidth]{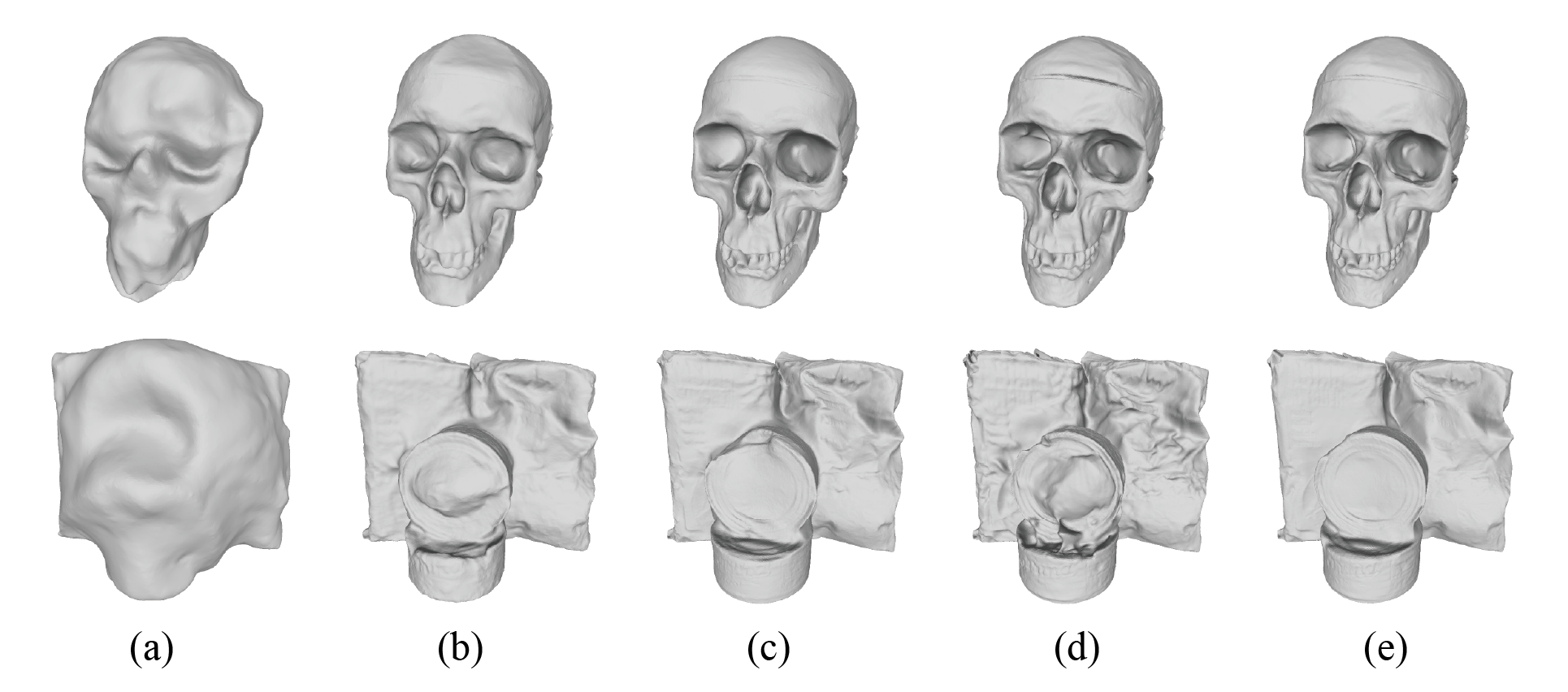}
  \caption{Qualitative results of ablation studies on DTU dataset. Symbols correspond to the settings in Table~\ref{tab1}.}
  \label{fig4}
\end{figure}

\textbf{Network Details.} For geometry modeling, inspired by DVR~\cite{60}, we implement $g_{\boldsymbol{\theta}}$ utilizing a fully connected ResNet block with 256-dimensional hidden features and a ReLU activation function. We set a multi-resolution hash grid with 16 levels, where the hash resolution ranges from 16 to 2048 and the feature dimension is 2. The output geometric feature $\boldsymbol{z}$ dimension is 64. For appearance modeling, we use a single MLP $h_{\boldsymbol{\sigma}}$ with 2 hidden layers, 64-dimensional hidden features, and ReLU activation. The spherical harmonics of the view direction are set to 3 degrees, and the last layer adopts a sigmoid activation function for the final colors. For geometric feature regularization, we use a small MLP $p_{\boldsymbol{\eta}}$ with 2 hidden layers, 256-dimensional hidden features, and a ReLU activation function.

\begin{figure}[t]
  \centering
  \includegraphics[width=\linewidth]{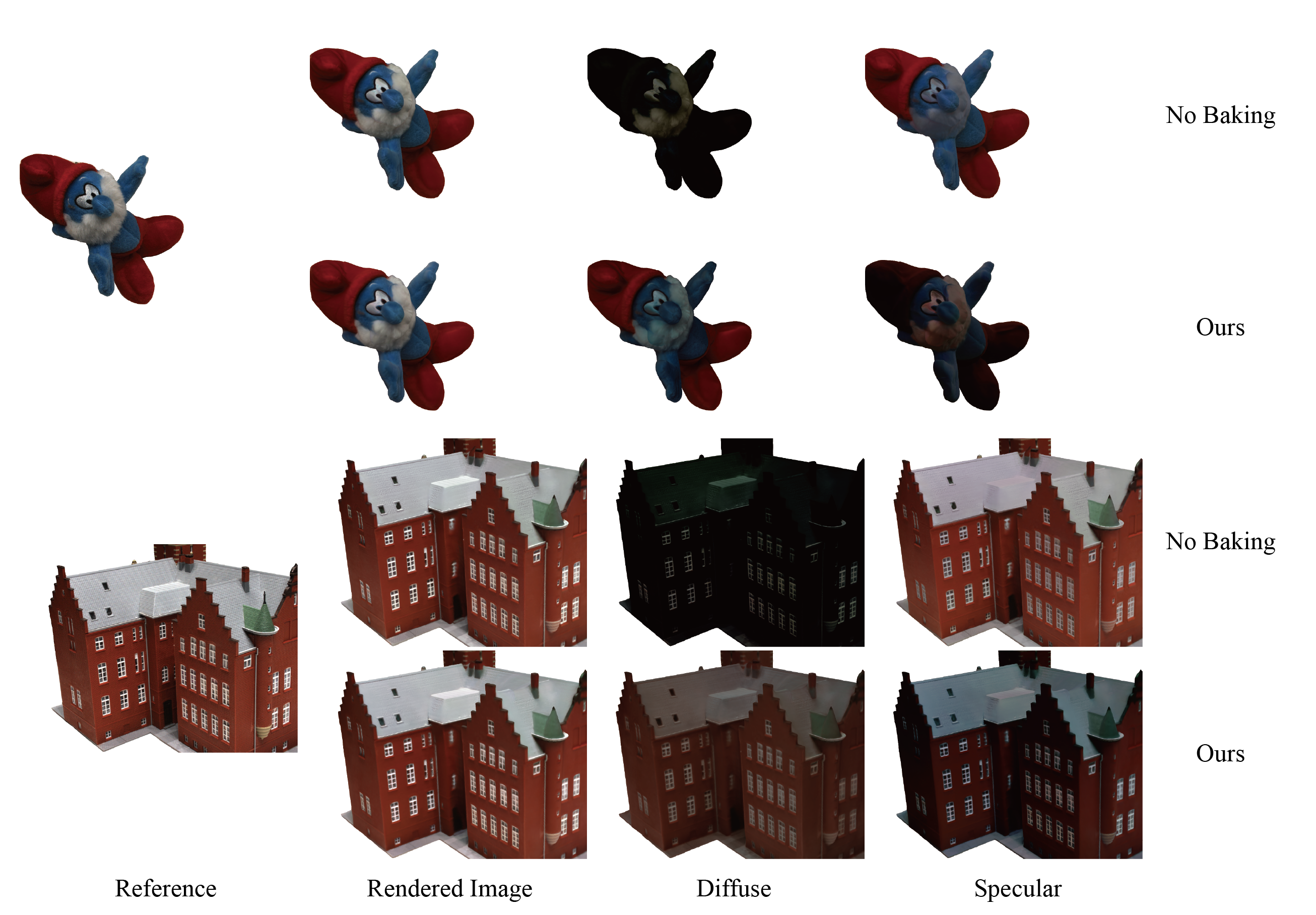}
  \caption{Results of view synthesis. Without the baking strategy, the output of the diffuse branch network would look dark, which is counterintuitive. In contrast, our method produces sensible results.}
  \label{fig5}
\end{figure}

\textbf{Datasets and Metrics.} All experiments are conducted on the DTU~\cite{61} and BlendedMVS~\cite{62} datasets. The DTU dataset is collected in a controlled laboratory setting and each scene contains 49 or 64 high-resolution images ($1600 \times 1200$), masks, camera poses, and ground truth 3D geometries. We use 15 challenging scans selected by IDR to test both quantitative and qualitative evaluations. The BlendedMVS dataset is captured in a natural setting and includes a wide range of scenes and objects. We further test the proposed method on 7 challenging scenes as NeuS, which has 31-143 images of $768 \times 576$ resolution. Using the official DTU~\cite{61} evaluation code, the chamfer distance (CD) between the reconstructed mesh and the GT mesh is employed as the evaluation metric for 3D reconstruction. The PSNR between the synthesized images and the reference images is used as the view synthesis evaluation metric. 

\subsection{Ablation Study}
As shown in Table~\ref{tab1} and Fig.~\ref{fig4}, we set five settings to verify the individual parts of the proposed method. The details of each design in Table~\ref{tab1} are summarized as follows:
\begin{itemize}
  \item Geometry MLP. Utilizes the neural geometry network $g_{\boldsymbol{\theta}}$ to compute the vertex deformation. Without this component, vertex offsets are directly treated as the learnable parameters like NDS~\cite{20}.
  \item Geometric Feature. Incorporates geometric features $z$ from $g_{\boldsymbol{\theta}}$ into subsequent rasterization and neural appearance modeling. Without this component, geometric features are removed from the rendering inputs.
  \item Diffuse Baking. Precomputes and stores the view-independent diffuse terms $c_{\boldsymbol{x}}^d$ at mesh vertices. Without this, an additional diffuse branch network, structurally similar to the specular branch $h_{\boldsymbol{\sigma}}$, is added to estimate the diffuse components.
  \item Apperance MLP. The neural shader used for rendering. This component is mandatory.
  \item $L_{feature}$. The geometric feature regularization term used to optimize geometric features. Without it, the loss function excludes this term.
\end{itemize}

Setting (a) serves as the baseline, using only the appearance MLP with vertex positions encoded by a multi-resolution hash grid. Both quantitative and qualitative results indicate limited performance. Setting (b) adds the geometric MLP and diffuse baking modules. This significantly improves learning and reconstruction quality, as evidenced by both metrics and visual results. Setting (c) further incorporates geometric features to enhance vertex positions. The results show notable improvements in PSNR, CD metrics, and shading quality, enabling more accurate and detailed reconstructions. Setting (d) introduces $L_{feature}$, representing the complete algorithm of this paper. This addition further optimizes geometric features, corrects errors, and enhances fine details. Quantitative results in Table~\ref{tab1} confirm this configuration achieves the best reconstruction and rendering performance.

Setting (e) removes diffuse baking and replaces it with a diffuse branch network. Although this setting has little effect on the CD metric, the rendered images exhibit significant distortions, with excessively dark diffuse components and dominant specular reflections, as shown in Fig.~\ref{fig5}. The PSNR decreases, and both training and rendering times increase, making this approach unsuitable for real-time applications. The results demonstrate that the proposed diffuse baking strategy effectively improves rendering and training efficiency while ensuring interpretable results.

Thus, the ablation study with different settings above have proved the effectiveness of each module of the algorithm in this paper.

\begin{table*}[t]
  \centering
  \caption{Quantitative comparisons of reconstruction (Chamfer Distance) and training time (TT) on DTU dataset.}
  \begin{tabular}{c|c|c|c|c|c|c|c|c}
    \hline
      & \multicolumn{3}{c|}{Neural Rendering-based Methods} & Voxel & \multicolumn{4}{c}{Mesh} \\
    \hline
     Scan & NeRF~\cite{7} & IDR~\cite{10} & NeuS~\cite{9} & DVGO~\cite{42} & FastMESH~\cite{21} & CPT-VR~\cite{23} & NDS~\cite{20} & Ours \\
    \hline
    
      24 & 1.83 & 1.63 & 0.83 & 1.83 & 0.65 &	0.55 & 4.24 &	1.71 \\
    
      37 & 2.39 & 1.87 & 0.98 & 1.74 & 1.48 & 0.72 & 5.25 & 3.24 \\
  
      40 & 1.79 & 0.63 & 0.56 & 1.70 & 0.57 &	0.35 & 1.30 & 0.81 \\

      55 & 0.66 & 0.48 & 0.37 & 1.53 & 0.40 & 0.38 & 0.53 & 0.52 \\

      63 & 1.79 & 1.04 & 1.13 & 1.91 & 1.48 & 0.85 & 2.47 & 2.29 \\

      65 & 1.44 & 0.79 & 0.59 & 1.91 & 0.77 & 0.59 & 1.22 & 0.76 \\

      69 & 1.50 & 0.77 & 0.60 & 1.77 & 0.56 & 0.54 & 1.35 & 0.95 \\

      83 & 1.20 & 1.33 & 1.45 & 2.60 & 0.86 & 0.81 & 1.59 & 1.69 \\

      97 & 1.96 & 1.16 & 0.95 & 2.08 & 0.84 & 0.83 & 2.77 & 1.17 \\

      105 & 1.27 & 0.76 & 0.78 & 1.79 & 0.94 & 0.67 & 1.15 & 0.91 \\

      106 &	1.44 & 0.67 &	0.52 & 1.76 &	0.72 & 0.53 &	1.02 & 0.85 \\

      110 &	2.61 & 0.90 &	1.43 & 2.12 &	0.81 & 0.59 & 3.18 & 3.45 \\

      114 &	1.04 & 0.42 &	0.36 & 1.60 &	0.52 & 0.32 &	0.62 & 0.40 \\

      118 &	1.13 & 0.51 &	0.45 & 1.80 &	0.49 & 0.38 & 1.65 & 0.69 \\

      122 &	0.99 & 0.53 & 0.45 & 1.58 &	0.54 & 0.37 &	0.91 & 0.68 \\
    \hline
      mean$\downarrow$ & 1.54 &	0.90 & 0.77 &	1.54 & 0.77 & 0.56 & 1.95 & 1.34 \\
    \hline
      TT (m)$\downarrow$ & 750 & 390 & 780 & 25 & 30 & 57.7 & 4.15 & 4.84 \\
    \hline
    \end{tabular}
  \label{tab2}
\end{table*}

\begin{table*}[t]
  \centering
  \caption{Quantitative comparisons of 2D view synthesis (PSNR) and rendering time (RT) on DTU dataset.}
  \begin{tabular}{c|c|c|c|c|c|c|c|c}
    \hline
      & \multicolumn{3}{c|}{Neural Rendering-based Methods} & Voxel & \multicolumn{4}{c}{Mesh} \\
    \hline
     Scan & NeRF~\cite{7} & IDR~\cite{10} & NeuS~\cite{9} & DVGO~\cite{42} & FastMESH~\cite{21} & CPT-VR~\cite{23} & NDS~\cite{20} & Ours \\
    \hline
    
      24 & 26.97 & 23.29 & 26.13 & 27.77 & 26.98 & 28.15 & 20.92 & 24.82 \\
    
      37 & 25.99 & 21.36 & 24.08 & 25.96 & 26.01 & 27.31 & 21.35 & 22.35 \\
  
      40 & 27.68 & 24.39 & 26.73 & 27.75 & 28.30 & 28.54 & 24.73 & 26.34 \\

      55 & 29.39 & 22.96 & 28.06 & 30.42 & 26.58 & 29.52 & 25.43 & 27.46 \\

      63 & 33.07 & 23.22 & 28.69 & 34.35 & 29.25 & 34.04 & 29.81 & 31.17 \\

      65 & 30.87 & 23.94 & 31.41 & 31.18 & 32.01 & 29.83 & 29.19 & 30.81 \\

      69 & 27.90 & 20.34 & 28.96 & 29.52 & 32.54 & 29.00 & 26.45 & 27.65 \\

      83 & 33.49 & 21.87 & 31.56 & 36.94 & 33.42 & 35.92 & 35.72 & 35.56 \\

      97 & 27.43 & 22.95 & 25.51 & 27.67 & 27.93 & 29.32 & 25.52 & 28.02 \\

      105 & 31.68 &	22.71 &	29.18 &	32.85 &	30.65 &	33.08 & 31.05 & 32.83 \\

      106 &	30.73 & 22.71 & 32.60 &	33.75 &	30.77 &	29.45 & 30.77 &	32.16 \\

      110 &	29.61 &	21.26 &	30.83 &	33.10 & 33.96 &	31.16 & 27.91 &	30.98 \\

      114 &	29.37 &	25.35 &	29.32 &	30.18 &	28.47 & 30.77 & 27.95 &	30.12 \\

      118 &	33.44 &	23.54 &	35.91 &	36.11 &	34.27 &	30.50 & 31.43 & 35.49 \\

      122 &	33.41 & 27.98 &	35.49 &	36.99 &	35.12 & 38.38 & 33.81 &	34.76 \\
    \hline
      mean$\uparrow$ & 30.07 & 23.20 & 29.63 & 31.64 & 30.42 & 31.00 & 28.14 & 30.04 \\
    \hline
      RT (s)$\downarrow$ & 30 &	30 & 60 &	0.75 & 0.10 &	0.057 &	0.117 &	0.023 \\
    \hline
    \end{tabular}
  \label{tab3}
\end{table*}

\begin{figure*}[t]
  \centering
  \includegraphics[width=\linewidth]{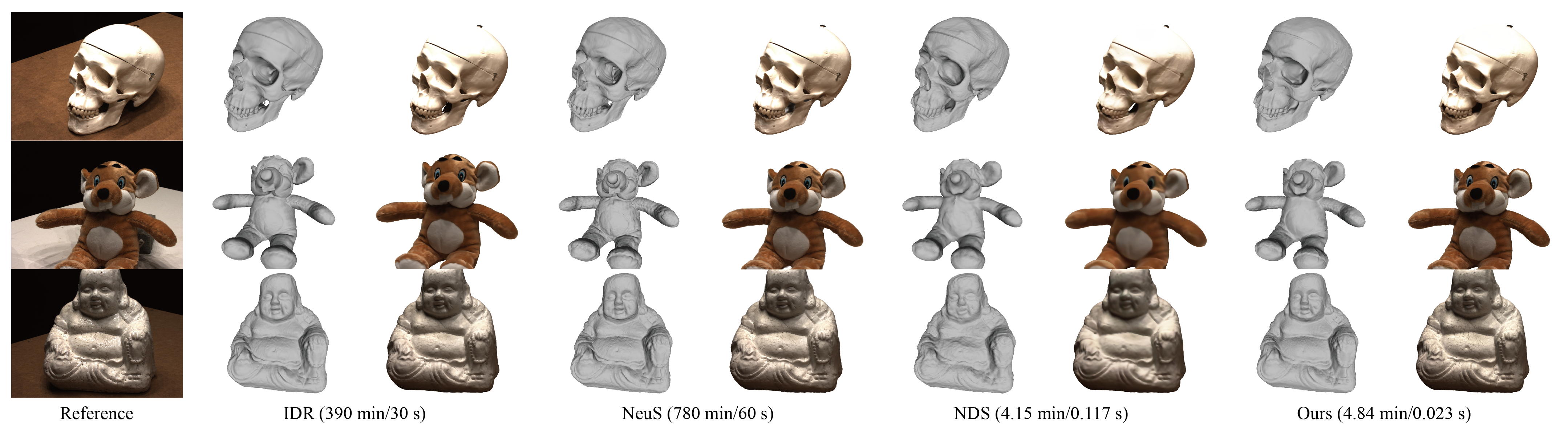}
  \caption{Qualitative comparisons on DTU dataset.The training time and rendering time of each method have been marked in the are denoted in brackets.}
  \label{fig6}
\end{figure*}

\subsection{3D Reconstruction and 2D View Synthesis}
The proposed method is compared with other state-of-the-art methods. We select representative methods of different types reviewed in Section 2 for comprehensive comparisons. Specifically, NeRF~\cite{7}, IDR~\cite{10}, and NeuS~\cite{9} represent neural rendering-based methods, DVGO~\cite{42} represents explicit voxel-based methods, NDS~\cite{20}, FastMESH~\cite{21}, and CPT-VR~\cite{23} represent explicit mesh-based methods. 

\begin{figure*}[t]
  \centering
  \includegraphics[width=\linewidth]{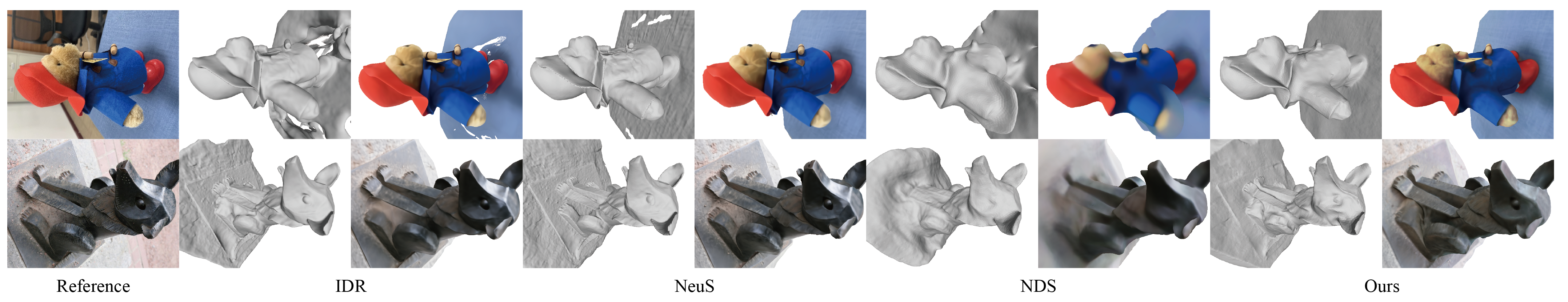}
  \caption{Qualitative results of objects and scenes on BlendedMVS dataset.}
  \label{fig7}
\end{figure*}

We evaluate the DTU dataset. Table~\ref{tab2} illustrates the quantitative reconstruction results and average training time. It can be found that methods with lower reconstruction errors, such as NeuS, CPT-VR, and FastMESH, usually require long training times, ranging from 30 to 780 minutes. In contrast, our method only takes about 5 minutes to train and achieves a moderate level of accuracy. It is worth mentioning that both FastMESH and CPT-VR utilize pre-trained MVS models as priors to improve mesh quality, which also adds additional constraints and pretraining costs. Our method does not use priors to ensure fast training and generalizability, and keeps the training time at the same level as the fastest method NDS, but greatly improves the accuracy. Fig.~\ref{fig6} shows the qualitative reconstruction comparison results. We use the official code and pre-trained models of other methods to obtain the reconstructed meshes. An extra Marching Cubes process is used to extract the meshes of the neural rendering-based methods. It can be observed that the mesh reconstructed using our method retains some details while maintaining a smooth surface. Fig.~\ref{fig7} shows more comparisons of reconstructed results on the BlendedMVS dataset. Our method performs well. The high-quality meshes generated by our method can be integrated into existing graphics workflows.

The quantitative comparisons of 2D view synthesis and the average rendering time are shown in Table~\ref{tab3}. The rendering time of the proposed method is the shortest and the rendered views achieve a high level of PSNR. Fig.~\ref{fig6} also shows the qualitative comparisons of 2D views. Our method can synthesize realistic images. The high efficiency and high quality of rendering are beneficial to some real-time applications.

\subsection{Mesh and Texture Editing}
We discuss several applications of the proposed method and demonstrate its wide applicability.

\begin{figure}[t]
  \centering
  \includegraphics[width=\linewidth]{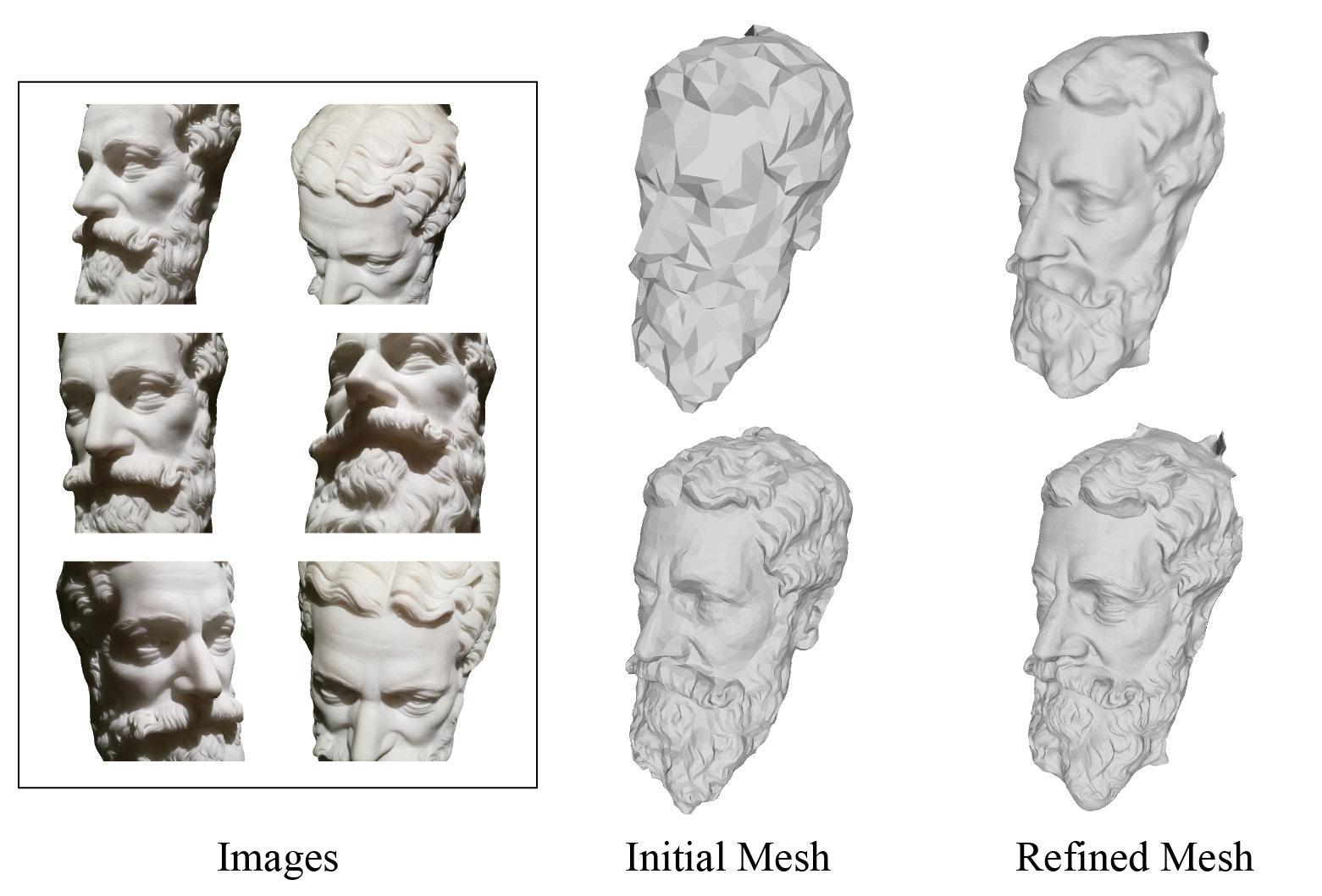}
  \caption{Mesh refinment of initial mesh with different resolutions.}
  \label{fig9}
\end{figure}
\begin{figure}[t]
  \centering
  \includegraphics[width=\linewidth]{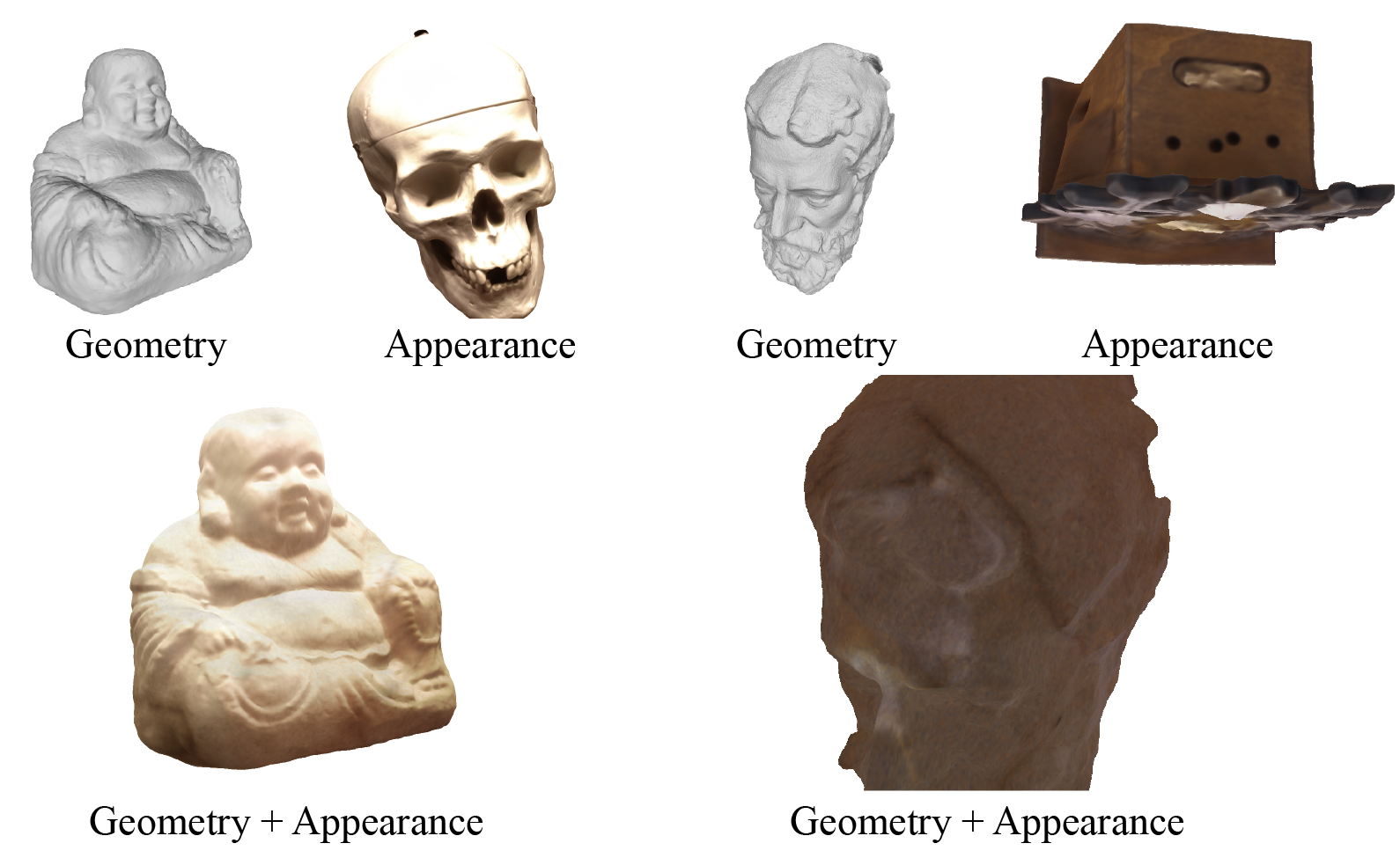}
  \caption{Different geometry and texture combination.}
  \label{fig10}
\end{figure}

\begin{figure*}[t]
  \centering
  \includegraphics[width=\linewidth]{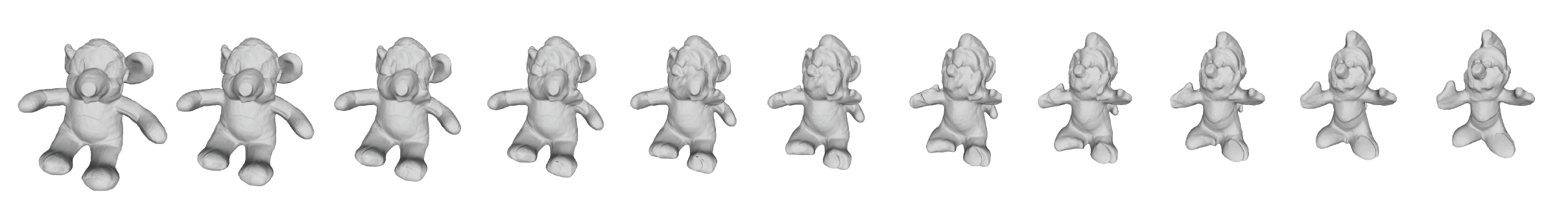}
  \caption{Mesh morphing between two objects (the far left and the far right) on DTU datasets.}
  \label{fig11}
\end{figure*}
\begin{figure}[t]
  \centering
  \includegraphics[width=\linewidth]{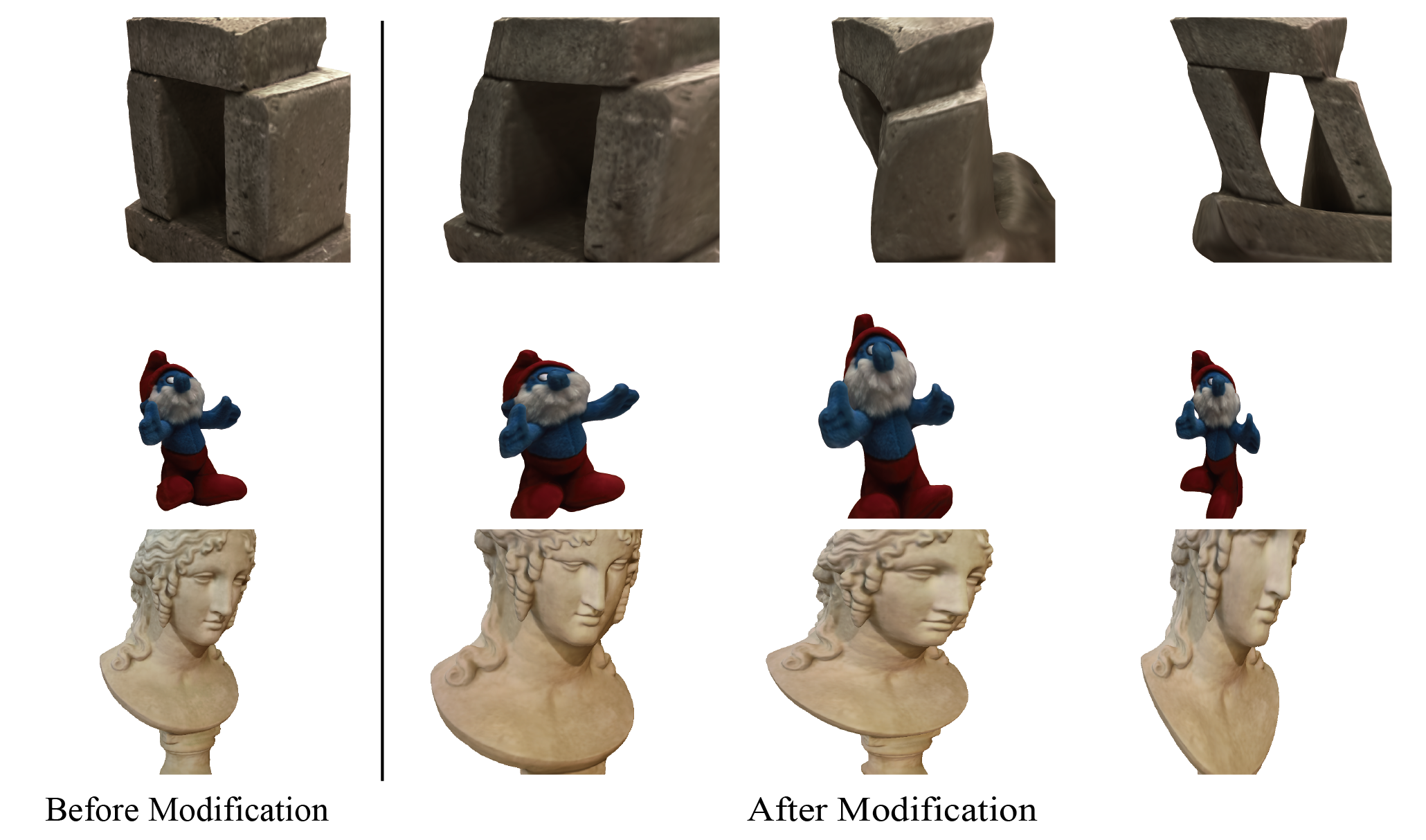}
  \caption{Novel view synthesis after mesh modification.}
  \label{fig12}
\end{figure}

\textbf{Mesh Refinement} Similar to NDS~\cite{20}, the proposed method can also be used as a post-processing step to refine coarse low-quality meshes, since they have in common that the initial mesh can be set to an arbitrary triangle mesh. As shown in Fig.~\ref{fig9}, we test initial meshes with different resolutions to demonstrate these applications. The initial meshes are obtained by simplifying the output of NeuS~\cite{9} to 1K (first row) and 10K vertices (second row), respectively. We apply three rounds of loop subdivisions in the first iteration, and the mesh resolution increases to about 32K and 320K vertices, respectively. After 2000 iterations, the final refined mesh is shown in the rightmost column of Fig.~\ref{fig9}. More fine details can be reconstructed after the refinement. It verifies that our method can further improve the mesh quality.

\textbf{Disentangling Geometry and Apperance.} In the proposed method, optimizing the geometry and appearance models separately helps to adjust the content of each part and combine, which expands the scope of applications, such as attribute transfer, mesh morphing, interpolation, and mesh-based modification. As shown in Fig.~\ref{fig10}, we can combine different geometries and appearances to obtain new objects and synthesize novel views. For attribute transfer, we train two scenes and save the geometry and appearance model parameters ($g_{\theta}$ and $h_{\sigma}$) of each scene, which can be replaced with each other. We test it on the DTU and BlendedMVS datasets, as shown in the left and right parts of Fig.~\ref{fig10}. It shows that the appearance can be successfully transferred to unseen geometries, and our method can synthesize nice novel views of unseen scenes.

Since the proposed method can obtain a topologically consistent final mesh using the same initial mesh, we can apply mesh morphing and interpolation between two shapes using two trained geometry models. Fig.~\ref{fig11} shows the full deformation process of two different objects in the DTU dataset. Some intermediate interpolated meshes are shown and the results are excellent.

Moreover, the proposed method supports mesh-based modification and adaptation. We deform the final mesh to obtain a new shape and synthesize novel views of the new mesh with the help of the trained appearance model. We test it on the DTU and BlendedMVS datasets, and the results of synthesizing novel views are shown in Fig.~\ref{fig12}. Realistic novel view results can be obtained, and even real-time animation can be achieved due to the fast rendering time. The above applications prove that our method is widely applicable.

\section{Conclusions}
In this paper, we propose a novel explicit mesh-based method for multi-view reconstruction and surface rendering. The disentangling of geometry and appearance modeling improves the learning capability. The proposed neural deformation field constrained by geometry regularization can inference more geometric information, such as robust features that are beneficial to reconstruction and rendering. The baking of view-invariant diffuse terms in our method improves rendering efficiency and helps real-time applications. Experiments demonstrate the effectiveness, great performance, and broad application prospects of our method.

\textbf{Superiority.} The proposed method provides a new solution for fast multi-view reconstruction and novel view synthesis. Without any priors, we can obtain high-quality mesh within 5 minutes in most scenes and can integrate it into existing graphics workflows. The rendering frequency of our method reaches 50 FPS, which is beneficial for real-time applications. Thanks to the high efficiency and the decoupling of geometry and appearance, our method contributes to many downstream works, including surface or texture editing and modification, and even real-time animations. 

\textbf{Limitations and Future Works.} The main limitations of our method are the same as most explicit mesh-based methods, that is, the topology of reconstructed mesh cannot be changed. This means that we need to carefully choose an initial mesh with a suitable topology at the beginning of reconstruction. In the future, we will investigate more flexible representations to address this issue. In this regard, we are excited about noticing some advanced works design mesh-based Gaussian Splatting representations for real-time deformation~\cite{63} and mesh rendering~\cite{64}, which may provide us with inspiration. In addition, we will consider a more effective strategy to learn more geometric information to further improve the accuracy of reconstruction while maintaining high efficiency.

\appendix

\subsection*{Availability of data and materials}

The data that support the findings of this study are available from the corresponding author upon reasonable request.

\subsection*{Author contributions}

\textbf{Qitong Zhang} is responsible for conceptualization, methodology, software, validation, formal analysis, investigation, resources, data curation, writing - original draft, writing - review \& editing and visualization. \textbf{Jieqing Feng} is responsible for conceptualization, methodology, validation, formal analysis, writing - review \& editing, supervision, project administration and funding acquisition.

\subsection*{Acknowledgements}
This work was jointly supported by the National Natural Science Foundation of China under Grant 62272408, Grant 61932018, and Grant 61732015.

\subsection*{Declaration of competing interest}

The authors have no competing interests to declare that are relevant to the content of this article.







\end{document}